\documentclass[sigconf]{acmart}
\AtBeginDocument{%
  }

\setcopyright{acmlicensed}
\copyrightyear{2018}
\acmYear{2018}
\acmDOI{XXXXXXX.XXXXXXX}
\acmConference[Conference acronym 'XX]{Make sure to enter the correct
  conference title from your rights confirmation email}{June 03--05,
  2018}{Woodstock, NY}
\acmISBN{978-1-4503-XXXX-X/2018/06}

\usepackage{amsmath}
\usepackage{graphicx}

\begin{document}

\title{Entity Representation Learning Through Onsite-Offsite Graph for Pinterest Ads}

\author{Jiayin Jin}
\email{jjin@pinterest.com}
\affiliation{%
  \institution{Pinterest}
  \country{USA}
}

\author{Erika Sun}
\email{zhengerikasun@pinterest.com}
\affiliation{%
  \institution{Pinterest}
  \country{USA}
}

\author{Zhimeng Pan}
\email{zpan@pinterest.com}
\affiliation{%
  \institution{Pinterest}
  \country{USA}
}

\author{Yang Tang}
\email{ytang@pinterest.com}
\affiliation{%
  \institution{Pinterest}
  \country{USA}
}

\author{Jiarui Feng}
\email{feng.jiarui@wustl.edu}
\affiliation{%
  \institution{Pinterest \& WUSTL}
  \country{USA}
}

\author{Kungang Li}
\email{kungangli@pinterest.com}
\affiliation{%
  \institution{Pinterest}
  \country{USA}
}

\author{Chongyuan Xiang}
\email{cxiang@pinterest.com}
\affiliation{%
  \institution{Pinterest}
  \country{USA}
}

\author{Jiacheng Li}
\email{jiachengli@pinterest.com}
\affiliation{%
  \institution{Pinterest}
  \country{USA}
}

\author{Runze Su}
\email{runzesu@pinterest.com}
\affiliation{%
  \institution{Pinterest}
  \country{USA}
}

\author{Siping Ji}
\email{siping@pinterest.com}
\affiliation{%
  \institution{Pinterest}
  \country{USA}
}

\author{Han Sun}
\email{hsun@pinterest.com}
\affiliation{%
  \institution{Pinterest}
  \country{USA}
}

\author{Ling Leng}
\email{lleng@pinterest.com}
\affiliation{%
  \institution{Pinterest}
  \country{USA}
}

\author{Prathibha Deshikachar}
\email{pdeshikachar@pinterest.com}
\affiliation{%
  \institution{Pinterest}
  \country{USA}
}

\renewcommand{\shortauthors}{Jin et al.}

\begin{abstract}
Graph Neural Networks (GNN) have been extensively applied to industry recommendation systems, as seen in models like GraphSage\cite{GraphSage}, TwHIM\cite{TwHIM}, LiGNN\cite{LiGNN} etc. In these works, graphs were constructed based on users' activities on the platforms, and various graph models were developed to effectively learn node embeddings. In addition to users' onsite activities, their offsite conversions are crucial for Ads models to capture their shopping interest. To better leverage offsite conversion data and explore the connection between onsite and offsite activities, we constructed a large-scale heterogeneous graph based on users' onsite ad interactions and opt-in offsite conversion activities. Furthermore, we introduced TransRA (TransR\cite{TransR} with Anchors), a novel Knowledge Graph Embedding (KGE) model, to more efficiently integrate graph embeddings into Ads ranking models. However, our Ads ranking models initially struggled to directly incorporate Knowledge Graph Embeddings (KGE), and only modest gains were observed during offline experiments. To address this challenge, we employed the Large ID Embedding Table technique and innovated an attention based KGE finetuning approach within the Ads ranking models. As a result, we observed a significant AUC lift in Click-Through Rate (CTR) and Conversion Rate (CVR) prediction models. Moreover, this framework has been deployed in Pinterest's Ads Engagement Model and contributed to $2.69\%$ CTR lift and $1.34\%$ CPC reduction. We believe the techniques presented in this paper can be leveraged by other large-scale industrial models.
\end{abstract}





\maketitle

\section{Introduction}
In recent years, Pinterest has become a popular destination for users, not only to explore new ideas and find inspiration but also to shop for new products, making it a fertile ground for advertisers to showcase their products and services. In response to the increasing demand for advertisements, Pinterest has developed a large-scale advertisement serving system. Understanding users' shopping intent and interest has become increasingly important for our ads system. Powered by Deep Learning, modern recommendation systems can incorporate hundreds or even thousands of signals to efficiently deliver personalized ads. Therefore, extracting features and signals from data originating from various sources has become essential for achieving success. In particular, conversions that occur outside of Pinterest, complementing onsite data, are instrumental to accurately capture users’ shopping intent. With this new opportunity, new challenges in feature engineering for offsite data have emerged. The first challenge is entity representations. Unlike onsite data, offsite entities often lack sufficient metadata coverage. Even when metadata is available, its format can vary across different advertisers. Secondly, discrepancies may exist between onsite and offsite data distributions, and establishing a mechanism to effectively integrate them could be beneficial. Motivated by these, we constructed a large-scale graph that integrates opt-in offsite conversion data with onsite ads data. Given the significant discrepancies between the metadata of onsite and offsite entities, a Knowledge Graph Embedding model, which does not rely on node features, is utilized to learn entity representations from the graph. 

The KGE models have been extensively studied in the literature, see for examples \cite{TransE, RESCAL, TransR, Complex}. To address concerns of scalability and training speed, we opted to employ a translation-based model. Inspired by the success of TwHIM\cite{TwHIM}, we initially experimented with the TransE\cite{TransE} model. Unfortunately, due to the inherent complexity and heterogeneity of our graph, the evaluation metrics of the TransE model were close to zero, indicating that the learned embeddings were not meaningful. While a more sophisticated model like TransR\cite{TransR} was able to generate meaningful embeddings, integrating the TransR model with downstream applications was challenging because different entity types resided in different spaces. To address these challenges, we introduced a novel TransRA model, TransR with Anchors, such that 1) designating one entity space as an anchor to which all other entity spaces are connected and 2) applying transformations only to non-anchor spaces. As a result, non-anchor spaces can be transformed into anchor space, improving the efficiency for downstream applications. 

Another key challenge was integrating KGE into ranking models, given that our ranking models are trained on tabular data with distributions that differ largely from graph data. Initial experiments using pretrained KGE within the ranking models resulted in only modest improvements. To address this, we implemented a strategy to load KGE into large embedding tables within the ranking models and fine-tune them during training. However, this approach did not yield gains in offline experiments as well. Finally, we innovated an attention-based KGE finetuning method. Specifically, we introduced a self-attention layer on top of all embeddings looked up from the KGE table to mimic graph interactions within the ranking model. Further explanations and discussion are provided in Section \ref{sec:integrate}. This approach led to substantial improvements in both online and offline experiments. This innovative approach not only addresses the challenges of integrating KGE into ranking models but also offers a novel perspective on enhancing large scale models by effectively leveraging diverse types and domains of data.

In summary, the contributions and innovations presented in this paper are as follows:  1. Constructing an large-scale onsite-offsite heterogeneous graph to learn latent representations for offsite entities; 2. Introducing a novel KGE model, TransRA, which effectively extracts information from complex heterogeneous graphs and is efficient in downstream applications; 3. Proposing an innovative approach of attention based fine-tuning large-scale KGE model within ranking models to achieve optimal performance. 

\section{Related Work}
Graph Neural Networks (GNNs), known for their capability to effectively propagate information through graphs and learn entity relations, have been widely applied in industry recommendation systems. Most of these applications utilize Graph Convolutional Neural Networks (GCNs), such as those described in \cite{GraphSage, ItemSage, LiGNN, GraphBid, FRANK, Alibaba}. In contrast, the TwHIM model \cite{TwHIM} employs a Knowledge Graph Embedding (KGE) approach. For additional references, readers are encouraged to consult the citations within those papers. Both approaches have their own advantages. GCN models aggregate features from neighboring nodes sampled by random walks, enabling them to leverage node features such as content features and metadata in addition to graph structures. This makes GCN models particularly useful for addressing cold start problems. In contrast, Knowledge Graph Embedding (KGE) models do not require nodes to have predefined features, offering greater flexibility in real-world applications. By learning node embeddings based solely on graph structures, KGE models excel at capturing global information. An intriguing future direction is to explore the combination of these two approaches. 

In addition to advancements in modeling, several GNN training frameworks have been developed to facilitate large-scale GNN training, including DGL, DGL-KE, PyTorch-BigGraph \cite{PyTorchBigGraph}, and PyG. Additionally, there are studies focused on graph design, such as those in \cite{Grale, Slaps}, with further references provided therein. Recently, there has been a growing trend in researching the integration of GNNs with large language models (LLMs); see, for example, \cite{LLaGA, TEA-GLM} and the references therein.

\section{Graph Construction and Model Training at Scale}
\subsection{Graph Construction}\label{sec: graph}
Our graph consisting of billions of nodes and edges is constructed from two data sources: onsite ad engagement and opt-in offsite conversion data. The primary goal of our work is to learn entity representations specifically for offsite data. We integrate offsite data with onsite data to address potential distribution discrepancies between them, thereby enhancing the efficiency of using our graph embeddings in downstream models. Given the strong relationship between ads and offsite conversions, including onsite ad engagement data is a natural choice. While existing works on heterogeneous graphs generally derive heterogeneity from different edge types within onsite data, our graph also introduces heterogeneity through an additional data source: offsite data.  

\begin{figure}
    \centering
    \includegraphics[width=\linewidth]{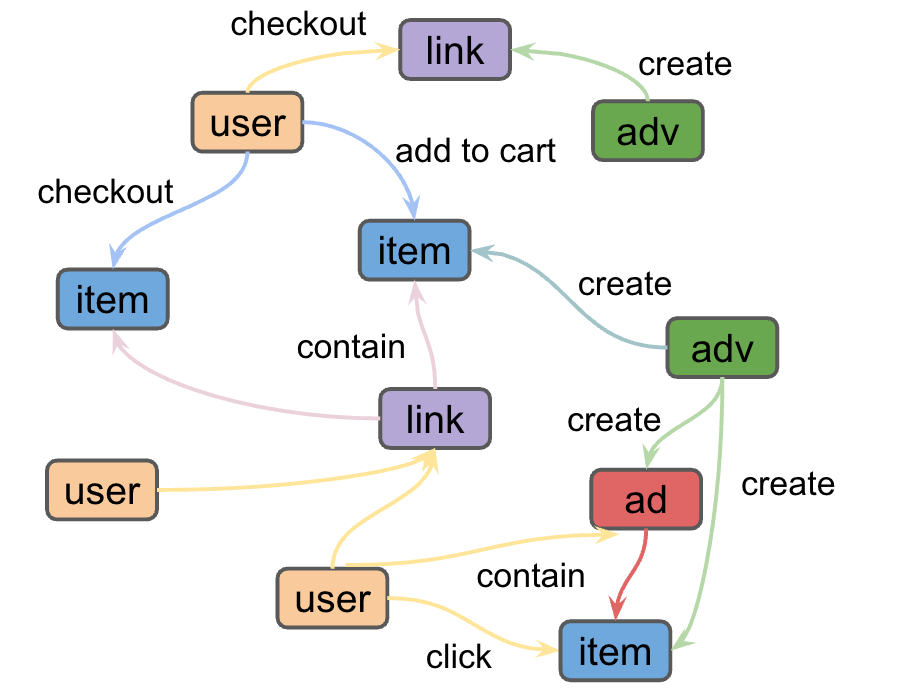}
    \caption{Illustration of onsite-offsite heterogeneous graph}
    \Description{Only present a subset of edge types here for the simplicity}
    \label{fig:graph}
\end{figure}

Our graph comprises five entity types: user, item, link, advertiser, and ad, along with more than ten edge types. These edge types can be categorized into three sets: onsite engagement edges include edges such as (user, click, ad) and (user, click, item), opt-in offsite conversion edges contains edges like (user, checkout, advertiser) and (user, checkout, item), parent-child edges consists of edges such as (advertiser, create, ad) and (ad, contain, item). See Figure \ref{fig:graph} for an illustration. 

\subsection{Background on KGE models}
KGE models learn node embeddings and relation transformations through the link prediction task. Each node is represented by an ID embedding, and each relation is associated with an operation to transform head and tail embeddings. Each edge is represented by $(h, r, t)$, where $h, r, t$ denote head node embedding, relation type, tail node embedding, respectively. For each edge $(h, r, t)$, a link prediction score $\mathcal{S}(h,r,t)$ is computed. For instance, in the TransE model\cite{TransE}, 
\begin{equation} \label{eq:transE}
\mathcal{S}(h,r,t) = d(h + T_r, t),
\end{equation} where $T_r$ is a translation embedding depending on $r$ and $d$ is a distance function such as $l^2$ distance, cosine similarity. 
In TransR model\cite{TransR}, 
\begin{equation} \label{eq:transR}
\mathcal{S}(h,r,t) = d(M_r*h + T_r, M_r*t),
\end{equation}
where $M_r$ is a projection matrix and $T_r$  a translation vector depending on the relation type. 

The training data of KGE models comprises positive and negative examples. Negative examples are typically generated through uniformly random sampling and in-batch sampling. The model is trained to distinguish positive edges from negative ones. Commonly used training loss functions are Sampled Softmax Loss\eqref{eq:softmax} and Marginal Ranking Loss \eqref{eq:ranking}. 
\begin{equation}\label{eq:softmax}
        \mathcal{L} = -\frac{1}{|P|}\sum_{(h,r,t)\in P}\log\frac{e^{\mathcal{S}(h,r,t)/\tau}}{e^{\mathcal{S}(h,r,t)/\tau} + \sum_{(h,r,t')\in N}e^{\mathcal{S}(h,r,t')/\tau}},
    \end{equation}
    where $P$, $N$ denote the set of positive examples and samples negative examples, respectively, $|p|$ represents the cardinality of the set $P$, and the parameter $\tau >0$ is the temperature.
\begin{equation}\label{eq:ranking}
        \mathcal{L} = \frac{1}{|P|\cdot |N|}\sum_{(h,r,t)\in P}\sum_{(h,r,t')\in N}\max(0, \lambda-(\mathcal{S}(h,r,t) - \mathcal{S}(h,r,t'))),
    \end{equation}
    where $\lambda > 0$ is the margin. 

\subsection{TransRA Model}\label{sec:model}
The central idea of the TransRA model is to designate one entity space as an anchor, such that all other entity spaces are connected to this anchor space through edges. Crucially, no transformations are applied to the anchor space for any edge type. This approach enables all other entity spaces to be transformed into the anchor space, enhancing the efficiency of integrating our graph embeddings with downstream models.

For convenience, the anchor space is always positioned on the head side. In TransRA, the score function $\mathcal{S}$ is defined as: 
\begin{equation}\label{eq:transra}
\mathcal{S}(h,r,t) =
    \begin{cases}
        \cos(h, M_{r}*t + T_{r}), & \text{if lhs is user}, \\
        \cos(M_{r}*h, M_{r}*t + T_{r}), & \text{if not}.
    \end{cases}
\end{equation}
Here $M_r$ is a projection matrix depending on edge type $r$, $T_r$ is a translation vector that also varies with the edge type. In the equation \eqref{eq:transra}, cosine similarity can be replaced by other distance functions such as the $l^2$-distance and so on. It is also worth mentioning that $M_r$ and $T_r$ are initialized as an identity matrix and a zero vector for any relation type $r$. Consequently, all entity embeddings are initialized in the same space, the model will determine how to allocate different entity spaces during training. 

Although the simple TransE model is capable of  translating all entity types into the same space, it struggles to capture complex relationships, such as n-to-m, particularly in heterogeneous graphs. The TransR model is adept at extracting information from complex heterogeneous graphs but it assigns different entity types to separate spaces, complicating their integration with downstream applications. The primary benefit of our TransRA model is its ability to effectively extract information from complex heterogeneous graphs while also integrating smoothly with downstream models. For our graph, the TransRA model outperforms the TransR model for edge types involving the anchor space, while achieving comparable performance for edge types that do not involve the anchor space. More details on the evaluation results can be found in Section \ref{sec:exp}. 

\subsection{Model Training}
We applied the TransRA model to the onsite-offsite heterogeneous graph described in Sec. \ref{sec: graph}. In this section, we provide details about model training. 
\begin{itemize}
    \item \textbf{Anchor Space:} Since user experience is our top priority, the user space is chosen as the anchor space. All other entities are connected to the user space through onsite engagement and opt-in offsite conversion activities. For instance, a user can be connected to an item via onsite click and opt-in offsite checkout actions.

    \item \textbf{Relation Type:} In our model training, relation types are distinct from edge types as defined in the graph. Relation types are determined solely by the head and tail entity types, and independent of the activity types. For example, the edge types (user, checkout, item) and (user, click, item) share the same relation type during training. This design choice is mainly for the simplifications of downstream use cases by eliminating the need to manage transformations for nodes shared by different edge types. 
    
    \item \textbf{Positive Examples:} Our model training process traverses all positive examples. During training, each edge type is assigned a dedicated dataloader, ensuring that every training batch includes all edge types. The proportion of each edge type in each training batch is determined by its data volume and significance. In particular, user engagement and conversion edges are allocated higher ratios.
    
    \item \textbf{Negative Sampling:} As in other KGE models, we employ two negative sampling methods: uniform sampling and in-batch sampling. Each edge type has a designated dataloader for loading uniformly sampled negatives. Note that, by uniform negatives we mean negative tails nodes. That is, for each positive $(h,r,t)$, we replace the tail node $t$ by some $t'$ from the uniform sampled negatives. In-batch negatives are generated by randomly permuting tail nodes. To balance model performance and training speed, each positive edge is paired with two uniformly sampled negatives and two in-batch negatives. 
    
    \item \textbf{Training Loss:} The Sampled Softmax Loss with temperature, as shown in Equation \eqref{eq:softmax}, is the most effective for our model. We set the temperature $\tau = 0.1$ during training to create a sharp distinction between positive and negative sets.
    On the other hand, the Marginal Ranking Loss \eqref{eq:ranking} and its variation \eqref{eq:ranking2}, does not perform well on our graph, resulting in nearly zero evaluation metrics. We tested both small and large margins. Even with a large margin, the distinction between positive and negative examples is insufficient, likely due to the large scale of our graph. More details can be found in Sec. \ref{sec:exp}.     
    \begin{align}\label{eq:ranking2}
    \mathcal{L} = \frac{1}{|P|} \sum_{(h,r,t) \in P} \max\Bigg(& 0, \lambda - \Bigg(\mathcal{S}(h,r,t) \nonumber \\
    & - \frac{1}{|N|} \sum_{(h,r,t') \in N} \mathcal{S}(h,r,t')\Bigg)\Bigg).
    \end{align}
    
    \item \textbf{Training Infrastructure:} 
    We implemented our in-house KGE model training pipeline which is compatible with Pinterest's Kubernetes based training platform.
    \begin{itemize} 
        \item \textbf{Graph Data Loading:}  Our graph data is stored as Parquet files partitioned by edge types. Each edge type contains hundreds of millions to billions of positive edges. During training, tens of streaming dataloaders operate in parallel, one for each edge type, to load batches of positive edges. The training batch, comprising around a hundred thousand edges, includes batches from all edge types.  For each non-anchor entity type, we randomly sample $2$ million nodes to create the uniform negative set. These negative examples are loaded during training to be paired with the corresponding positive edges.

        \item \textbf{Distributed Model Parallel:} Our model utilizes large ID embedding tables, where each node is represented by a trainable ID embedding. Given the large scale of our graph, these embedding tables contain hundreds of millions to billions of rows. As a result, Distributed Data Parallel cannot be used, as a single GPU does not have sufficient memory to accommodate the entire embedding tables. Storing the embedding tables on the CPU is another option, however, it incurs excessive CPU-GPU communication costs, which significantly slow down training. For training efficiency, we employ TorchRec's EmbeddingBagCollection and DistributedModelParallel modules to shard these large embedding tables across multiple GPUs. It takes approximately 20 hours for one P4d machine to train for $100k$ iterations, with each batch consisting of around $50k$ edges

        \item \textbf{Mixed Precision Training:} To conserve GPU memory, we use FP16 precision for the embedding tables, while maintaining FP32 precision for the transformation matrices and translation vectors. A single P4D instance can support the training of a KGE model with around $500$ million embeddings, each with 256 dimensions. It is possible to trade off embedding dimensions for the number of embeddings if needed. Additionally, our setup supports multi-node training, allowing the use of multiple P4D instances as the graph scales up.
    \end{itemize}
\end{itemize}

\section{Integrating KGE into Ads Ranking Models}\label{sec:integrate}
In this section, we present our journey of integrating KGE model into Pinterest’s ad ranking models. Further experimental results are detailed in Section \ref{eq:ranking}.

\subsection{Utilizing Pretrained Embeddings} 
We began by building a daily incremental training workflow to continuously update our KGE model with the latest data. The newly generated embeddings were integrated into the ranking models on a daily basis. This method, however, led to only a marginal AUC improvement of $0.03\%$ in our CVR model, which was considered statistically insignificant.

\subsection{Direct Finetuning} 
Given the hypothesis that pretrained graph embeddings may not be fully leveraged by ad ranking models trained on tabular data, we explored the approach of fine-tuning the KGE within the ranking models. We introduced a large KGE table into the ranking models, initializing it and associated node transformations with the pretrained KGE model snapshot. The transformed embeddings were used as input to the ranking models, allowing both the embedding table and transformations to be jointly fine-tuned. This approach, however, yielded neutral results on offline metrics as well. Our hypothesis is that the pre-encoded graph information diminished during finetuning.

\subsection{Attention-Based Finetuning} 
\begin{figure*}[h]
    \centering
    \includegraphics[scale=0.7]{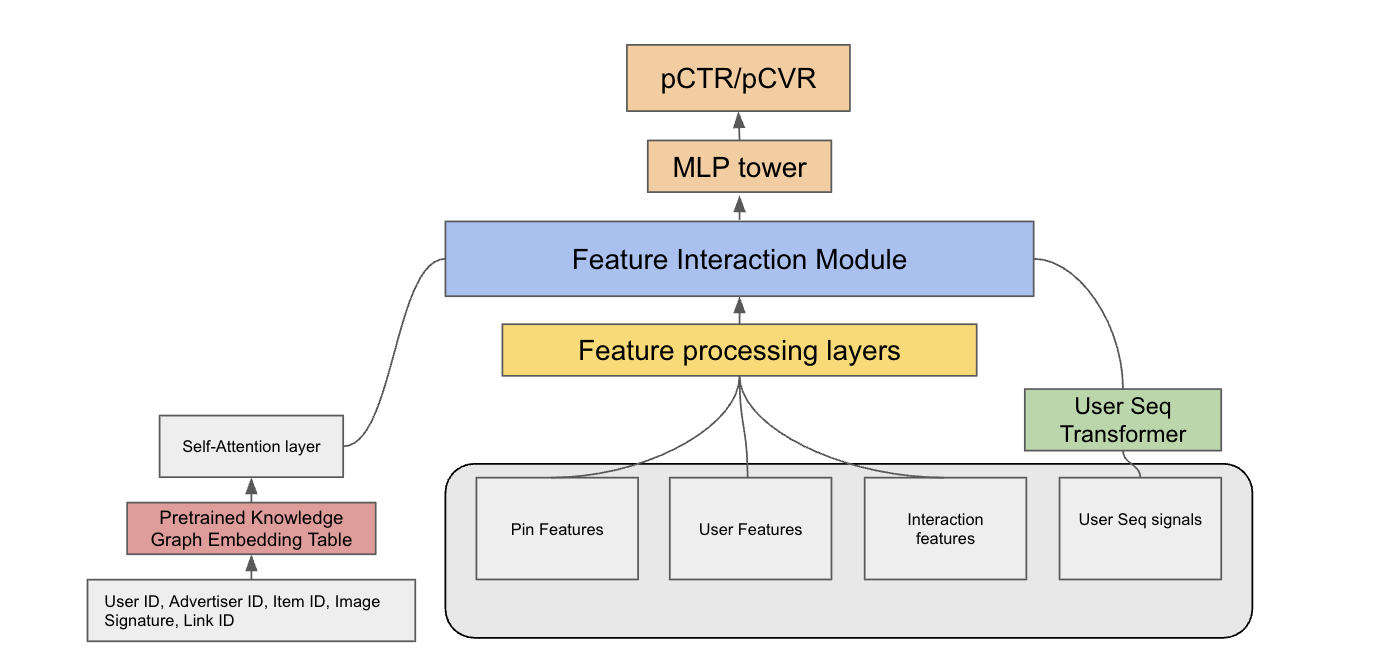}
    \caption{Attention Based Finetuning}
    \Description{Integrate with Ranking Model}
    \label{fig:ranking}
\end{figure*}

Finally, we innovated an attention-based finetuning method, as illustrated in Figure \ref{fig:ranking}, designed to better capture graph relations of different entities, similar to those in KGE model training. More specifically, before feeding embeddings into the ranking models, we passed them through a self-attention layer to mimic the graph interactions as in the KGE model. The heuristics underlying this approach and the analogy to the KGE model training are summarized below in Table \ref{t:heuristics}. As shown in Table \ref{t:finetune}, this method led to significant performance improvements.

\begin{table}[ht]
    \centering
    \caption{Heuristics of Attention-Based Finetuning}
    \begin{tabular}{|p{3.5cm}|p{3.5cm}|}
        \hline
        \textbf{KGE Model Training} & \textbf{KGE Table Finetuning} \\
        \hline
        Sample positive and negative ID pairs & The training data includes both positive and negative examples, where pairs within positive examples are treated as positive pairs, and pairs within negative examples are treated as negative pairs \\
        \hline
        Compute scores for each pair & Pairwise scores are computed by the self-attention layer \\
        \hline
        Train with a loss function to differentiate positive and negative pairs & All scores are fed into the ranking model and trained together with other features through a ranking loss to differentiate positive examples from negative ones \\
        \hline
    \end{tabular}
    \label{t:heuristics}
\end{table}

\section{Experimental Results}\label{sec:exp}
\subsection{KGE Model Performance}
\begin{itemize}
    \item \textbf{Evaluation Set:} We hold out $500k$ positives for each edge type as our evaluation set. Each positive edge $(h, r, t)$ is ranked against $30k$ pairs $(h, r, t')$ with uniformly sampled $t'$. 
    \item \textbf{Evaluation Metrics:} The main metrics are $$Recall@k = \frac{\# positives\; ranked\; within\; top\; k}{\# positives}$$
    with $k = 10$ and $100$.
    \item \textbf{Evaluation Results:} For simplicity in presentation, we provide evaluation metrics for only a subset of selected edge types. As illustrated in Table \ref{t:recall10} and Table \ref{t:recall100}, our TransRA model performed comparably to TransR for edge types without the anchor space. Furthermore, it outperformed TransR on edge types involving anchor space. Given the complexity of our heterogeneous graph, the TransE model failed to produce meaningful embeddings. 
  
As demonstrated in Table \ref{t:emb_dim10} and Table \ref{t:emb_dim100}, we observed that embedding dimensions significantly affect the performance of our graph. This impact is particularly pronounced for complex edge types such as user-item and user-ad, but it is less noticeable for simpler edges like ad contains item. Therefore, trading embedding dimension for the scale of the graph may not be advisable if we aim for best performance. Alternatively, enabling multi-node training could be a more effective approach when scaling up the graph.

\end{itemize}

\begin{table}[ht]
    \centering
    \caption{Recall@10 of Selected Edge Types with 256 Dimensional Embedding }
    \begin{tabular}{|c|c|c|c|}
        \hline
        & \multicolumn{3}{|c|}{Recall@10} \\
        \hline
        Edge Type & TransRA & TransR & TransE  \\
        \hline
        User Checkout Item & \textbf{0.812} & 0.81 & 1.2e-4 \\
        \hline
        User Click Ad & \textbf{0.586} & 0.54 & 0.1  \\
        \hline
        Ad contain Item & 0.994 & \textbf{0.996} & 0.02 \\
        \hline
    \end{tabular}
    \label{t:recall10}
\end{table}

\begin{table}[ht]
    \centering
    \caption{Recall@100 of Selected Edge Types with 256 Dimensional Embedding }
    \begin{tabular}{|c|c|c|c|}
        \hline
         & \multicolumn{3}{|c|}{Recall@100} \\
        \hline
        Edge Type & TransRA & TransR & TransE \\
        \hline
        User Checkout Item & 0.98 & \textbf{0.99} & 1.5e-3 \\
        \hline
        User Click Ad & \textbf{0.827} & 0.806 & 0.25 \\
        \hline
        Ad contain Item & \textbf{0.999} & \textbf{0.999} & 0.05 \\
        \hline
    \end{tabular}
    \label{t:recall100}
\end{table}

\begin{table}[ht]
    \centering
    \caption{Recall@10 for Different Embedding Dimension}
    \begin{tabular}{|c|c|c|}
        \hline
        & \multicolumn{2}{|c|}{Recall@10}  \\
        \hline
        Embedding Dimension & 256 & 64 \\
        \hline
        User Checkout Item & 0.812 & 0.463 \\
        \hline
        User Click Ad & 0.586 & 0.28  \\
        \hline
        Ad contain Item & 0.994 & 0.969 \\
        \hline
    \end{tabular}
    \label{t:emb_dim10}
\end{table}
\begin{table}[ht]
    \centering
    \caption{Recall@100 for Different Embedding Dimension}
    \begin{tabular}{|c|c|c|}
        \hline
        & \multicolumn{2}{|c|}{Recall@100} \\
        \hline
        Embedding Dimension & 256 & 64 \\
        \hline
        User Checkout Item & 0.98 & 0.807 \\
        \hline
        User Click Ad & 0.827 & 0.513 \\
        \hline
        Ad contain Item & 0.999 & 0.999 \\
        \hline
    \end{tabular}
    \label{t:emb_dim100}
\end{table}

\subsection{Offline Evaluation Metrics in Ads Ranking Models}\label{exp:ranking}
As described in Section \ref{sec:integrate}, we evaluated three approaches for integrating our KGE model with the Ads ranking models: (1) directly utilizing daily refreshed pretrained embeddings, (2) direct finetuning of KGE within ranking models, and (3) attention-based finetuning. We conducted experiments for each method for Ads ranking models. Below, we summarize our key findings:
\begin{itemize}
\item \textbf{TransR vs. TransRA in the CVR model:} Since offline experiments for the CTR model require more training data and backfilling pretrained KGE embeddings is resource-intensive, we began by testing pretrained embeddings in the CVR model. We tested both TransR and TransRA models in the CTR model, though neither of them produced significant gains, the TransRA model outperformed the TransR model as shown in Table \ref{t:pretrain}, which demonstrated the effectiveness of our TransRA model in the downstream applications. 

\item \textbf{Direct Finetuning vs. Attention Based Finetuning within the CTR model}: Next, we evaluated both finetuning approaches in the CTR model. As demonstrated in Table \ref{t:finetune}, the direct finetuning approach yielded neutral results, whereas the attention-based finetuning method resulted in significant gains.

\item \textbf{Attention Based Finetuning within the CVR model}: Building on the success of attention-based finetuning in the CTR model, we extended this method to the CVR model. As shown in Table \ref{t:cvr}, this approach yielded substantial offline gains.

\item \textbf{Training KGE Tables from Scratch}: We also evaluated the attention-based large KGE table model without loading pretrained embeddings. The results showed a neutral $+0.01\%$ AUC change in both CTR and CVR models, demonstrating the effectiveness of our onsite-offsite KGE model.

\begin{table}[ht]
\centering
\caption{Offline Evaluation Metrics of Pretrained KGE in the CVR model}
\begin{tabular}{|c|c|c|}
     \hline
      & AUC & PR\_AUC \\
    \hline
    TransR & +0.01\% & -2.83\% \\
    TransRA & +0.03\% & +0.57\% \\
    \hline
\end{tabular}
\label{t:pretrain}
\end{table}

\begin{table}[ht]
\centering
\caption{Comparison of different finetuning methods for the CTR model}
\begin{tabular}{|c|c|c|}
     \hline
      & AUC & PR\_AUC \\
    \hline
    Direct  & +0.003\% & +0.007\% \\
    Attention & +0.09\% & +0.28\% \\
    \hline
\end{tabular}
\label{t:finetune}
\end{table} 

\begin{table}[ht]
\centering
\caption{Offline Evaluation Metrics of KGE Table in CVR models}
\begin{tabular}{|c|c|c|}
     \hline
      & AUC & PR\_AUC \\
    \hline
    Click Through CVR & +0.52\% & +17.1\% \\
    View Through CVR & +0.41\% & +14.5\% \\
    \hline
\end{tabular}
\label{t:cvr}
\end{table} 
\end{itemize}

\subsection{Online Experimental Metrics}
We conducted an online bid segmented experiment test to assess the efficacy of the CTR model incorporating the KGE table. As illustrated in Table \ref{t:online}, the integration of the KGE model resulted in statistically significant improvements across key online metrics. Notably, the Cost-Per-Click (CPC) metric, which is a central performance indicator for our CTR ads, was improved substantially. Notably, the model with KGE table has been \textbf{deployed in the production}. In light of these encouraging results, we anticipate analogous gains within our Conversion Rate (CVR) model, and an online experiment is scheduled. 
\begin{table}[ht]
    \centering
    \caption{Online Metrics of KGE Table in the CTR model}
    \begin{tabular}{|c|c|c|}
         \hline
         Max-bid Revenue & CTR & CPC \\
        \hline
         +2.29\% & +2.6\% & -1.34\%  \\
        \hline
    \end{tabular}
    \label{t:online}
    \end{table} 
\section{Conclusion}
In this paper, we have introduced several key innovations: 
\begin{itemize}
    \item We constructed a large-scale heterogeneous graph consisting of users' onsite ad interactions and opt-in offsite conversion activities, providing a comprehensive view of user behavior.
    \item We developed TransRA, a novel KGE model that is able to extract information from complex heterogeneous graphs and efficiently integrate with ranking models. 
    \item We proposed an innovative attention based finetuning approach to finetune large KGE tables within ranking models, effectively addressing the distribution discrepancy that arises when integrating pretrained KGE directly into these models.
\end{itemize}
Both our offline and online experiments demonstrated that these methodologies significantly enhance ad performance. We believe our approach holds the potential for widespread industry applications.

\bibliographystyle{ACM-Reference-Format}
\bibliography{reference}

\end{document}